\documentclass[11pt]{article}

\usepackage[preprint]{acl}

\usepackage{times}
\usepackage{latexsym}

\usepackage[T1]{fontenc}
\usepackage[utf8]{inputenc}

\usepackage{microtype}

\usepackage{inconsolata}

\usepackage{graphicx}
\usepackage{booktabs}
\usepackage[most]{tcolorbox}
\usepackage{makecell}  
\usepackage{multirow}

\title{When to Stop Asking: Factor Coverage Bias and LLM Determinability Judgments with Underspecified Information in Medicine}

\author{
  \textbf{Yusuke Watanabe\textsuperscript{1,2,$\ast$}},
  \textbf{Yohei Kobashi\textsuperscript{3}},
  \textbf{Taisei Tosaki\textsuperscript{1}},
  \textbf{Takeshi Kojima\textsuperscript{3}},\\
  \textbf{Yusuke Iwasawa\textsuperscript{3}},
  \textbf{Yasushi Okuno\textsuperscript{1}},
  \textbf{Yutaka Matsuo\textsuperscript{3}}
\\\\
  \textsuperscript{1}Kyoto University, Department of Biomedical Data Intelligence\\
  \textsuperscript{2}Kyoto University, Department of Cardiovascular Medicine\\
  \textsuperscript{3}The University of Tokyo\\
  \textsuperscript{$\ast$} \texttt{usukewatanabe@gmail.com}
}

\begin{document}
\maketitle

\begin{abstract}
Clinical decision-making often begins with underspecified information and requires not only gathering missing information but also knowing when to stop. This calls for judging determinability (whether the available information is sufficient for a decision) by considering whether missing information could change the conclusion, not simply the amount of available decision-relevant information. Reliance on this amount as a cue for determinability, which we term \textit{factor coverage bias}, could undermine the trustworthiness of large language models (LLMs) through premature conclusions or excessive information gathering, such as unnecessary invasive tests. Yet whether LLMs exhibit this bias remains insufficiently studied. To address this gap, we construct a controlled pairwise benchmark in which LLMs identify the determinable case within each determinable--undeterminable pair. We add a decision-relevant factor to either side while keeping each case's determinability unchanged, and similarly add irrelevant text to distinguish this bias from context-length effects. Adding a relevant factor made LLMs more likely to select the augmented case as determinable even when it remained undeterminable, an effect that persisted with increased reasoning effort, while adding irrelevant text made them less likely to select the augmented case, an effect that reasoning substantially reduced. These findings identify factor coverage bias as a failure mode distinct from context-length effects and a potential risk to LLM trustworthiness. Our work provides an evaluation framework for LLM trustworthiness under underspecification.
\footnote{Our benchmark and code are available at \url{https://github.com/yusukewatanabe1208/when_to_stop_asking}.}
\end{abstract}

\begin{figure}[t]
    \centering
    \includegraphics[width=\linewidth]{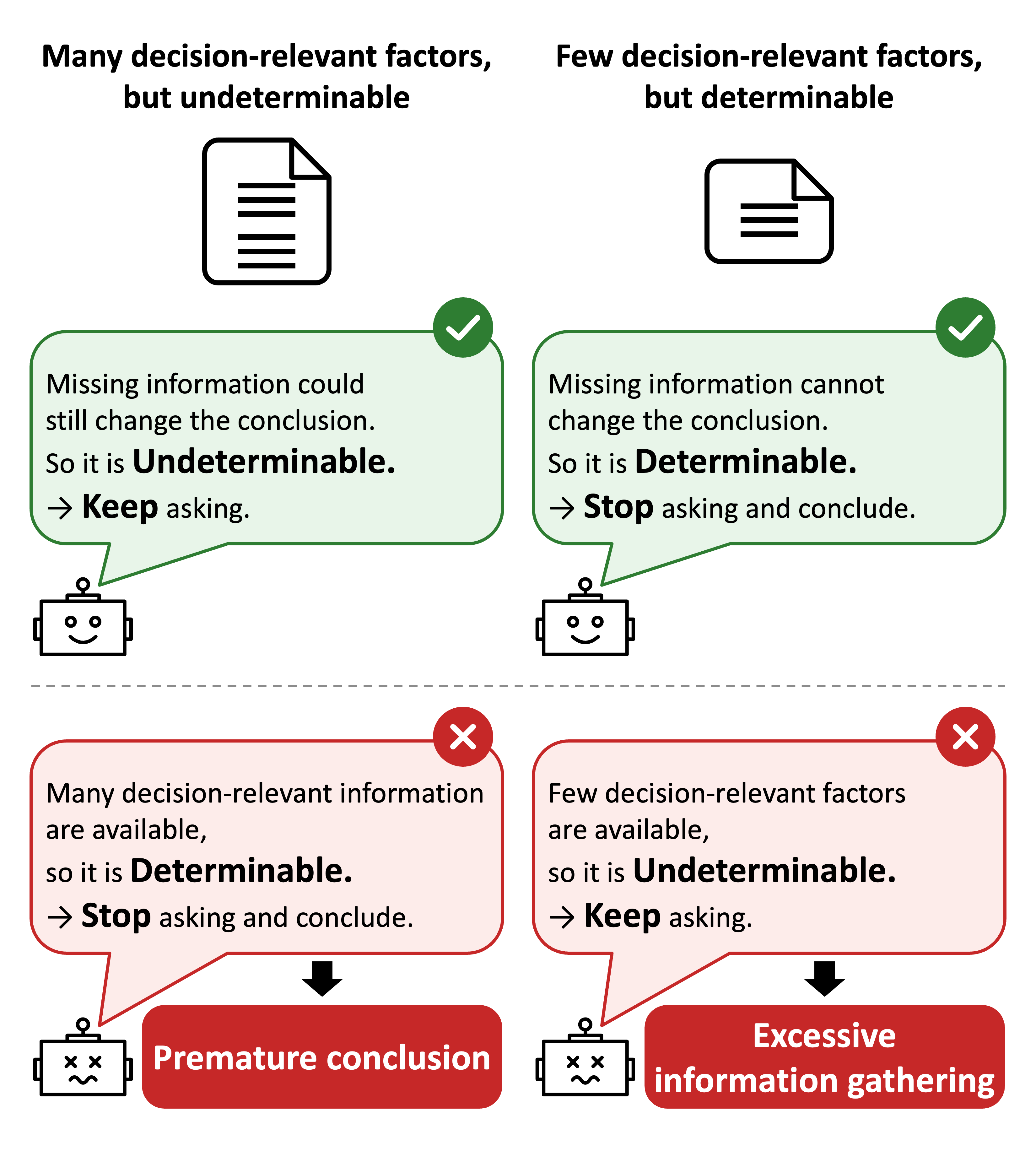}
    \caption{Concept of factor coverage bias. Determinability depends on whether missing information could change the conclusion. An LLM exhibiting factor coverage bias tends to use the amount of available decision-relevant information as a cue for determinability. This can lead to errors in deciding when to stop gathering information: premature conclusions when many factors are available but still insufficient, and excessive information gathering when few factors already suffice.}

    \label{fig:concept}
\end{figure}

\section{Introduction}

Recent LLMs have reached near-ceiling performance on benchmarks based on medical licensing examinations \citep{tordjman_Comparative_2025,liu_Textbooklevel_2026}.
The challenge is shifting from knowledge to reasoning as experts do in practice \citep{li_MediQ_2024,watanabe_ClinDetBench_2026}.
Examinations, however, typically provide the information needed for a decision, whereas clinical practice often involves underspecified information \citep{hersh_Caveats_2013,tobin-schnittger_Improving_2018}, limiting what these examinations can reveal about reasoning under such conditions \citep{hager_Evaluation_2024}. In clinical practice, experts tend to focus information gathering on what is needed for a decision, while trainees tend to collect more unnecessary information \citep{bowen_Educational_2006,mcquade_Characteristics_2024}.

When inputs are underspecified, as in patient-facing history taking and decision support for physicians outside their specialty, LLMs need not only to gather missing information but also to know when to stop, avoiding premature conclusions and excessive information gathering. In clinical practice, unnecessary questioning in emergencies or unnecessary invasive testing can harm patients \citep{grady_Less_2010}, so gathering every potentially relevant detail is not necessarily appropriate even if the final conclusion is correct. Prior work examines abstention under insufficient information and gathering additional information \citep{machcha_Knowing_2026a,li_MediQ_2024}, but stopping once information is sufficient has received less attention. \citet{watanabe_ClinDetBench_2026} defined \textit{determinability} by whether missing information could change the conclusion and demonstrated failures in LLM determinability judgments, without fully examining which input cues drive them. Failures in determinability judgment may undermine LLM trustworthiness beyond medicine, including coding assistance, legal consultation, and customer support, where robust responses to underspecified inputs require clarifying user intent and gathering necessary information, yet asking about every potentially relevant detail burdens users and reduces efficiency \citep{fang_ClarifyCodeBench_2026a,vincent_InsufficiencyBench_2026,mass_Conversational_2022}.

LLM biases, such as sycophancy \citep{sharma_Understanding_2024a}, can undermine trustworthiness by reducing robustness. 
LLMs are also sensitive to context length and distraction by irrelevant content \citep{shi_Large_2023a,amiraz_Distracting_2025,du_Context_2025}. Inputs grow through both relevant information and irrelevant text, especially during retrieval or multi-turn interactions.
However, more relevant information does not necessarily imply determinability. Few factors can suffice, while many can still be insufficient. LLMs may still use the amount of available decision-relevant information as a cue for determinability, a tendency we term \textit{factor coverage bias} (Figure~\ref{fig:concept}). This bias could lead to premature conclusions or excessive information gathering and may require different mitigation from context-length effects, because filtering irrelevant text would not prevent reliance on the amount of relevant information. Adding relevant information also lengthens the context, but whether factor coverage biases determinability judgments independently of context length remains insufficiently studied.

To address these gaps, we construct a controlled pairwise benchmark of 1,080 pairs based on clinical scoring systems to examine the bias clearly.
Each pair contains a determinable and an undeterminable case, both with missing information, and LLMs identify which is determinable.
We test for factor coverage bias by adding a decision-relevant factor to either side while keeping each case's determinability unchanged and measuring how the judgment shifts.
We similarly add irrelevant text of comparable length to either side to distinguish this bias from context-length effects.
Across recent open-weight and proprietary LLMs, adding a relevant factor shifts judgments toward determinable even when the augmented case remains undeterminable, an effect that persists with increased reasoning effort, while adding irrelevant text shifts judgments toward undeterminable, an effect that reasoning substantially reduces.
These differences in direction and response to reasoning identify factor coverage bias as a failure mode distinct from context-length effects.
Our main contributions are as follows.

\begin{itemize}
\item We define and demonstrate factor coverage bias, the tendency to use the amount of available decision-relevant information as a cue for determinability, as a potential risk to LLM trustworthiness.
\item We distinguish factor coverage bias from context-length effects through their different directions and responses to reasoning, informing future efforts to develop trustworthy LLMs.
\item We release a controlled pairwise benchmark for evaluating factor coverage bias in LLMs.
\end{itemize}

\section{Related Work}

\paragraph{LLM behavior under underspecified input.}
Underspecified inputs are a practical challenge in LLM deployment \citep{yang_What_2026,laban_LLMs_2025}. Missing information in the input is distinct from gaps in the model's knowledge \citep{ren_Dont_2026a}. Prior work has evaluated LLM responses to underspecified inputs by examining whether they ask clarifying questions \citep{zhang_CLAMBER_2024,dong_PRACTIQ_2025,li_QuestBench_2025,luo_ClarifyMTBench_2025,fang_ClarifyCodeBench_2026a}, abstain from answering \citep{kirichenko_AbstentionBench_2026}, or flag what is missing \citep{vincent_InsufficiencyBench_2026}. Few evaluate whether the conclusion could still change given what is missing \citep{watanabe_ClinDetBench_2026}, and systematic biases in this judgment remain insufficiently understood. Our work characterizes these biases.

\paragraph{Heuristics in LLM judgment.}
Input-driven biases undermine the robustness and trustworthiness of LLMs. Known examples include sycophancy \citep{sharma_Understanding_2024a} and evaluators' preference for longer responses \citep{zheng_Judging_2023,dubois_LengthControlled_2024}. Adding irrelevant text or increasing context length can reduce accuracy \citep{shi_Large_2023a,amiraz_Distracting_2025,du_Context_2025}, but the direction of the resulting answer shifts remains underexplored. Increasing supporting evidence shifts answers toward the supported side \citep{xie_Adaptive_2024,naphade_Rational_2026}, but these studies do not examine information sufficiency judgments. In these studies, the amount of relevant information and context length also change together, leaving their effects difficult to distinguish, although they may require different mitigation. We address these gaps by comparing additions of relevant information and irrelevant text while holding determinability fixed and examining how judgments shift.

\paragraph{LLM benchmarks in medicine.}
Medical benchmarks for LLMs have largely assumed complete information, as in those derived from licensing examinations \citep{jin_What_2021,pal_MedMCQA_2022b}. Work on incomplete information has evaluated abstention \citep{machcha_Knowing_2026a,cocchieri_LLMs_2026}, requests for additional information \citep{arora_HealthBench_2025}, and interactive information gathering \citep{li_MediQ_2024,schmidgall_AgentClinic_2026}. ClinDetBench \citep{watanabe_ClinDetBench_2026} evaluates determinability judgment directly, but because its items are constructed by progressively removing information from complete cases, factor coverage and determinability are correlated by construction, so the benchmark cannot detect whether a model responds to factor coverage rather than to determinability. We manipulate the two independently and evaluate this failure mode.

\section{Methodology}
\label{sec:method}

\subsection{Motivation}
We adopt a pairwise format because it controls for each model's propensity to abstain, which confounds single-case judgments, and because pairwise comparison amplifies superficial cues \citep{jeong_Comparative_2025a}, making bias easier to detect. Case presentations are constructed from clinical scoring systems. The aim is not to evaluate the ability to apply scoring rules; rather, because the judgment is determined by item values and a threshold, the correct label can be constructed uniquely, which yields a robust experimental setting.

\subsection{Preliminary: Clinical Scoring Systems}
Clinical scoring systems are widely used rule-based tools that assign points to clinical findings and guide decisions such as treatment initiation or risk stratification according to whether the total score reaches a threshold. A representative example, CHADS$_2$, assesses the risk of stroke in patients with atrial fibrillation. It assigns one point each to congestive heart failure (C), hypertension (H), age $\ge 75$ (A), and diabetes (D), and two points to a history of stroke or transient ischemic attack (S$_2$), and classifies a total score $\ge 2$ (i.e., threshold $\theta = 2$) as high risk (Table~\ref{tab:a1_chads2_items}). We use 15 scoring systems that are recommended in clinical guidelines and have explicit thresholds, each with 4--9 scored items (Table~\ref{tab:a2_scoring_systems}).

% ---------------------------------------------------------------------
% Table A.2. CHADS2 scored items and points (the chunks are Table A.4)
% ---------------------------------------------------------------------
\begin{table}[!t]
  \centering
  \small
  \caption{CHADS$_2$ scoring system. Total scores range from 0 to 6; we use a threshold of $\theta = 2$.}
  \label{tab:a1_chads2_items}
  \begin{tabular}{lc}
    \toprule
    Item & Points if positive \\
    \midrule
    Congestive heart failure & 1 \\
    Hypertension            & 1 \\
    Age $\ge 75$            & 1 \\
    Diabetes mellitus       & 1 \\
    Prior stroke or TIA      & 2 \\
    \bottomrule
  \end{tabular}
\end{table}

\subsection{Definition of Determinability}

When the values of some scored items are missing, the exact total score cannot be computed, but the judgment can still be determined if all possible total scores fall on the same side of the threshold. Let $S_{\min}$ and $S_{\max}$ denote the minimum and maximum possible total scores given the available information, obtained by assigning each item with a missing value its minimum and maximum possible points, respectively. With threshold $\theta$, a case is determinable if $S_{\min} \ge \theta$ or $S_{\max} < \theta$, and undeterminable if $S_{\min} < \theta \le S_{\max}$.

For example, in CHADS$_2$ ($\theta = 2$), a patient aged 80 with a history of heart failure has a total score of at least 2 regardless of the remaining missing items and is therefore determinable. In contrast, a patient with hypertension but without heart failure or diabetes, whose age and stroke history are missing, has a possible total score ranging from 1 to 4 and is therefore undeterminable. Determinability thus depends on whether missing information could change the conclusion, rather than the amount of available relevant information.

\subsection{Task Formulation}
\begin{figure}[t]
    \centering
    \includegraphics[width=0.90\linewidth]{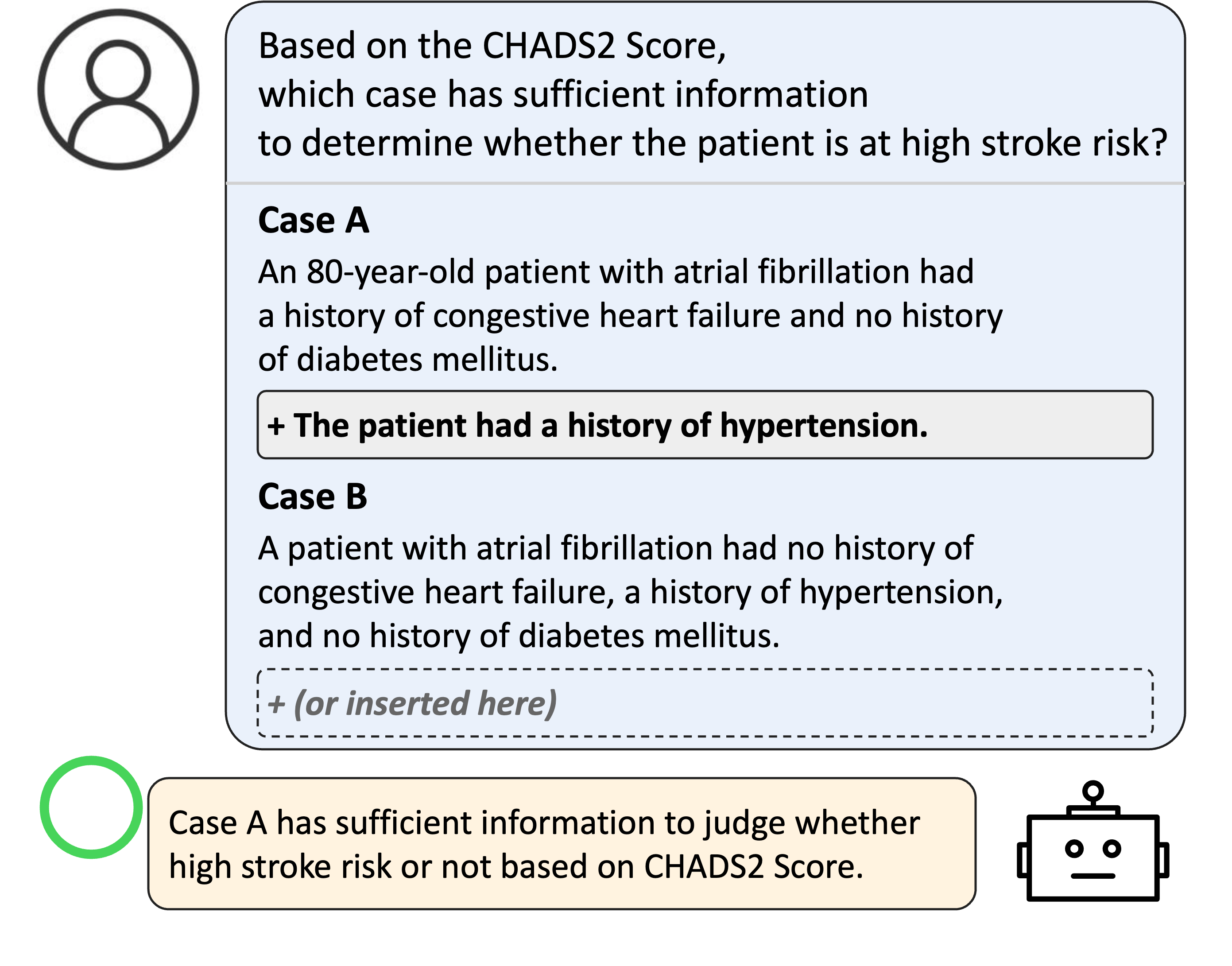}
    \caption{Overview of our pairwise benchmark. LLMs identify which case is determinable. We add a decision-relevant factor to either the determinable or the undeterminable case. We similarly add irrelevant text to either case, yielding four conditions in total. Each addition preserves both cases' determinability. We compare the resulting judgment shifts to distinguish factor coverage bias from context-length effects.}
    \label{fig:overview}
\end{figure}

Figure~\ref{fig:overview} provides an overview of our benchmark. 
Each pair contains a determinable and an undeterminable case constructed from the same clinical scoring system. The model identifies, in a forced binary choice, which case contains sufficient information to settle the judgment under that system. Both cases have missing scored items. One is determinable, meaning that the judgment is the same for all possible values of the missing items, while the other is undeterminable, meaning that the judgment depends on those values.

\subsection{Dataset Construction}
\paragraph{Materials.}
Case presentations are constructed from pre-written text chunks that explicitly indicate whether each documented scored item is positive (scoring) or negative (non-scoring). Table~\ref{tab:a4_scored_item_chunks} illustrates these chunks using CHADS$_2$. Irrelevant sentences describe medically plausible details, such as drug allergies, current medications, or family history, that do not affect the score (Table~\ref{tab:a8_irrelevant_chunks}). A board-certified internal medicine physician checked the chunks against item definitions and verified that irrelevant sentences neither directly nor implicitly indicated any scored item. For example, irrelevant sentences were checked for mentions of antidiabetic drugs in scoring systems that include diabetes. Determinability labels are assigned automatically by computing the possible total score range ($S_{\min}$, $S_{\max}$) and comparing it with the threshold. The prompt template and examples of the actual prompts are provided in Appendix~\ref{sec:app_prompt}.

\paragraph{Baseline pairs.}
A baseline pair consists of a determinable case and an undeterminable case with the same number of missing items and comparable length to minimize differences in information amount and context length at baseline. Eight pairs were sampled uniformly at random within each scoring system, yielding 120 baseline pairs, with correct answers balanced between A and B. Each baseline pair is used in nine conditions, including baseline, yielding 1,080 pairs in total ($120 \times 9$). Baseline cases average 41 tokens, with a mean prompt length of 140 tokens (Table~\ref{tab:a5_baseline_by_score}).

\paragraph{Adding a relevant factor.}
To test for factor coverage bias, we examine whether adding relevant information shifts judgments toward the augmented case as determinable. A relevant factor is realized as a positive or negative chunk for a previously missing scored item (Table~\ref{tab:a4_scored_item_chunks}), which we add to either case of a baseline pair. Baseline pairs are constructed so that the undeterminable case remains undeterminable after the addition, while the determinable case also retains its label. This increases both the amount of available relevant information and the context length without changing either case's determinability. The addition averages 8 tokens.

\paragraph{Adding irrelevant text.}
To distinguish factor coverage bias from context-length effects, we similarly add irrelevant sentences (Table~\ref{tab:a8_irrelevant_chunks}) to either case without changing its determinability. The length-matched condition adds one sentence with the same token count as the added relevant factor for the same scoring system (mean 8 tokens). To examine larger length increases, the 2$\times$ and 4$\times$ conditions extend the case to approximately twice and four times its original length, adding 43 and 127 tokens on average, respectively. Token counts by condition are reported in Table~\ref{tab:a7_condition_stats}.
\begin{table*}[t]
\centering
\caption{Error rates. Det/Und denote the determinable/undeterminable case to which text is added; values in parentheses are changes from baseline. Irrelevant text is length-matched to the added relevant factor. Reasoning indicates the setting used, with \texttt{max} for GPT-5.5 and \texttt{high} for Opus 4.5 representing the highest reasoning levels evaluated; for the Qwen models, \texttt{none}/\texttt{on} correspond to thinking off/on.}
\label{tab:relevant_vs_irrelevant}
\footnotesize
\setlength{\tabcolsep}{4pt}
\renewcommand{\arraystretch}{0.92}
\setlength{\aboverulesep}{0.2ex}
\setlength{\belowrulesep}{0.3ex}
\begin{tabular*}{\textwidth}{@{\extracolsep{\fill}}llccccc@{}}
\toprule
 & & & \multicolumn{2}{c}{Add relevant factor} & \multicolumn{2}{c}{Add irrelevant text (matched)} \\
\cmidrule(lr){4-5}\cmidrule(lr){6-7}
Model & Reasoning & Baseline & Det & Und & Det & Und \\
\midrule
\multirow{2}{*}{GPT-5.5} & none & 0.342 & 0.233 ($-$0.108) & 0.525 ($+$0.183) & 0.475 ($+$0.133) & 0.267 ($-$0.075) \\
 & max & 0.017 & 0.017 ($\pm$0.000) & 0.050 ($+$0.033) & 0.025 ($+$0.008) & 0.025 ($+$0.008) \\
\addlinespace[2pt]
\multirow{2}{*}{Opus 4.5} & none & 0.425 & 0.358 ($-$0.067) & 0.525 ($+$0.100) & 0.558 ($+$0.133) & 0.392 ($-$0.033) \\
 & high & 0.125 & 0.042 ($-$0.083) & 0.233 ($+$0.108) & 0.142 ($+$0.017) & 0.133 ($+$0.008) \\
\addlinespace[2pt]
\multirow{2}{*}{Qwen3.5-27B} & none & 0.342 & 0.225 ($-$0.117) & 0.492 ($+$0.150) & 0.542 ($+$0.200) & 0.183 ($-$0.158) \\
 & on & 0.100 & 0.050 ($-$0.050) & 0.242 ($+$0.142) & 0.208 ($+$0.108) & 0.108 ($+$0.008) \\
\addlinespace[2pt]
\multirow{2}{*}{Qwen3.5-9B} & none & 0.375 & 0.350 ($-$0.025) & 0.433 ($+$0.058) & 0.617 ($+$0.242) & 0.275 ($-$0.100) \\
 & on & 0.167 & 0.100 ($-$0.067) & 0.233 ($+$0.067) & 0.192 ($+$0.025) & 0.217 ($+$0.050) \\
\addlinespace[2pt]
\multirow{2}{*}{Qwen3.6-35B-A3B} & none & 0.375 & 0.367 ($-$0.008) & 0.508 ($+$0.133) & 0.500 ($+$0.125) & 0.308 ($-$0.067) \\
 & on & 0.200 & 0.108 ($-$0.092) & 0.292 ($+$0.092) & 0.233 ($+$0.033) & 0.192 ($-$0.008) \\
\bottomrule
\end{tabular*}
\end{table*}

\subsection{Evaluation}

\label{sec:evaluation}
For each model and each condition, we use the error rate against the correct label (1 $-$ accuracy) as the metric.
The effect of an addition is quantified by \textbf{attraction}. Because the correct answer is always the determinable side, a tendency to select the side to which text was added increases errors when the addition is made to the undeterminable side and decreases errors when it is made to the determinable side. We therefore define $\mathrm{attraction} = \mathrm{Err}_{\mathrm{Und}} - \mathrm{Err}_{\mathrm{Det}}$, where $\mathrm{Err}_{\mathrm{Und}}$ and $\mathrm{Err}_{\mathrm{Det}}$ are the error rates when the text is added to the undeterminable and determinable side, respectively. Attraction is positive when the judgment is drawn toward the added side, negative when the added side is avoided, and near 0 when an addition merely degrades both sides equally (non-directional noise). We obtain 95\% confidence intervals (CIs) by a pair-level paired bootstrap (10,000 resamples, percentile method).

\section{Experiments}
\label{sec:experiments}

\subsection{Experimental Setup}
We evaluated GPT-5.5, Claude Opus 4.5, and three open-weight models, Qwen3.5-27B, Qwen3.5-9B, and Qwen3.6-35B-A3B, through their providers' APIs, with the Qwen models accessed via OpenRouter. Reasoning effort was varied through the thinking level for Opus 4.5 (\texttt{none}/\texttt{low}/\texttt{medium}/\texttt{high}), the reasoning-effort parameter for GPT-5.5 (\texttt{none}/\texttt{low}/\texttt{high}/\texttt{max}), and thinking off/on for the Qwen models (reported as \texttt{none}/\texttt{on}). Temperature was left at the provider default, and each instance was queried once.
The prompt was a single zero-shot template containing the scoring system, the decision target, and the two cases (Figure~\ref{fig:a1_prompt_template}); no judgment procedure was supplied, because our question is whether models make the judgment autonomously.

\subsection{Overall Error Rates}
Table~\ref{tab:relevant_vs_irrelevant} reports error rates and changes from baseline. Enabling reasoning reduced error rates across all five models and all conditions shown, with baseline error rates falling from 0.342--0.425 to 0.017--0.200 (see also Figure~\ref{fig:scaling_reasoning_tokens} in the appendix). Adding a relevant factor to the determinable case reduced or left unchanged the error rate, whereas adding it to the undeterminable case increased the error rate in every model and reasoning setting. With reasoning enabled, adding a relevant factor to the undeterminable case produced the highest error rate for every model among the conditions shown. Without reasoning, irrelevant text showed the opposite pattern, increasing errors when added to the determinable case and decreasing errors when added to the undeterminable case; these changes were generally smaller with reasoning. The next section quantifies these directional shifts as attraction. Results including intermediate reasoning levels and the 2$\times$ and 4$\times$ conditions are provided in Table~\ref{tab:a9_error_rate_all}.

\subsection{Factor Coverage Bias Is Distinct from Context-Length Effects}
\begin{figure}[t]
    \centering
    \includegraphics[width=\linewidth]{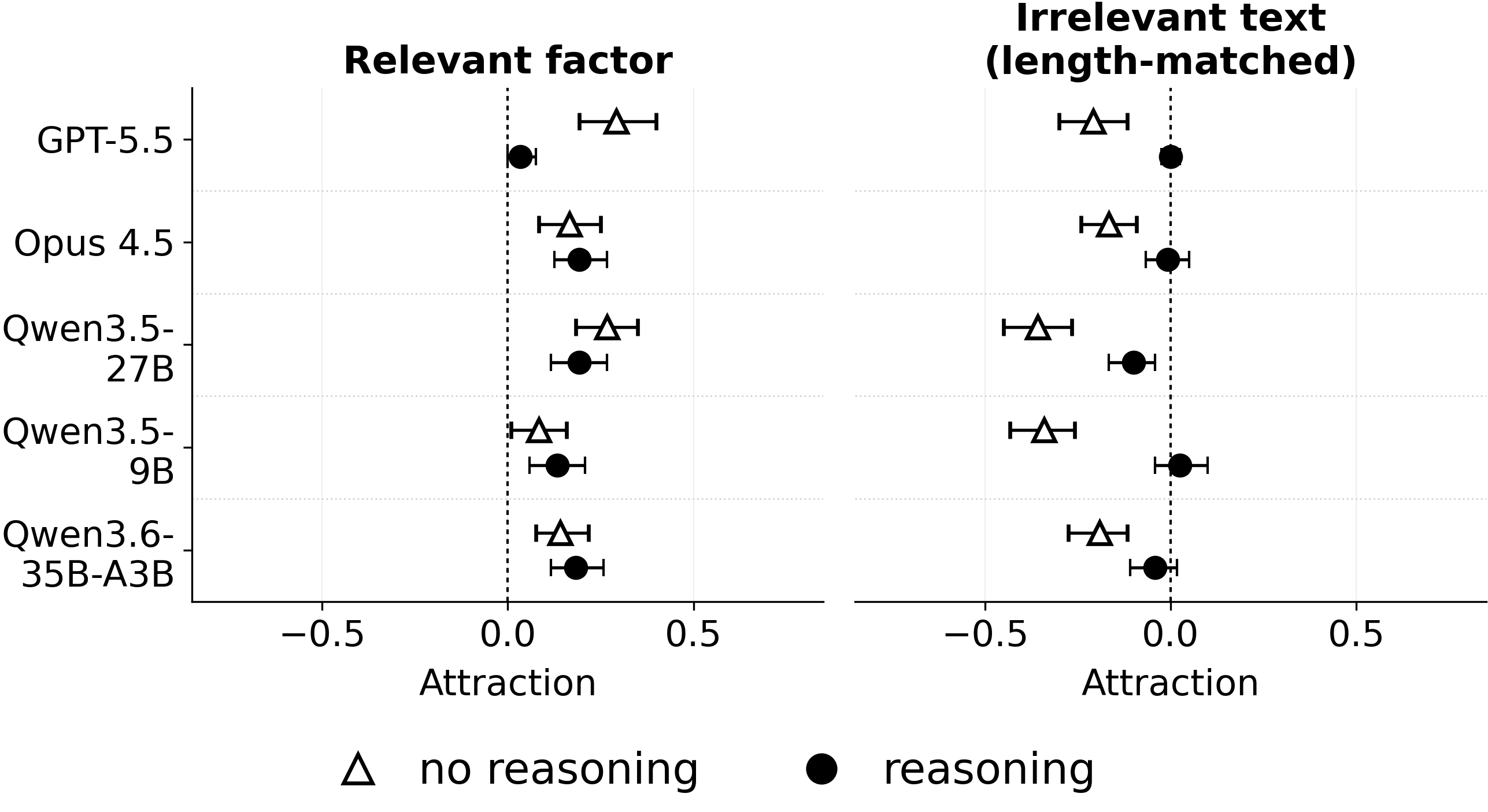}
    \caption{Attraction with 95\% CIs for the addition of one relevant factor and of length-matched irrelevant text. Positive values indicate that the judgment was drawn toward the added side.}
    \label{fig:attraction}
\end{figure}

Figure~\ref{fig:attraction} shows the attraction (Section~\ref{sec:evaluation}) and its 95\% CI for each model with reasoning off and on, for the addition of one relevant factor (left) and the addition of length-matched irrelevant text (right); \texttt{on} denotes the highest level available (\texttt{max} for GPT-5.5 and \texttt{high} for Opus 4.5).

\begin{figure}[t]
    \centering
    \includegraphics[width=\linewidth]{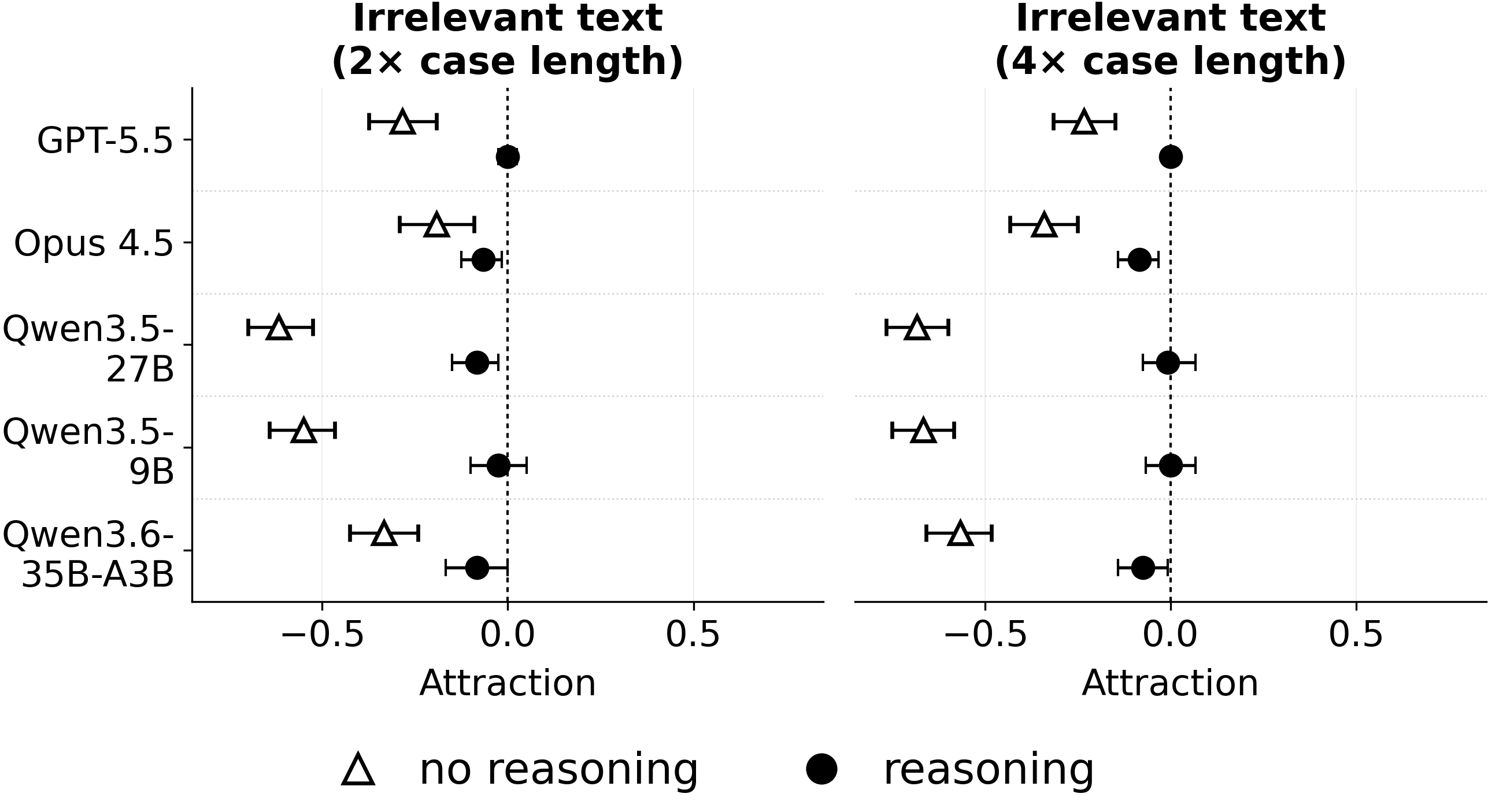}
    \caption{Attraction with 95\% CIs for irrelevant text at 2$\times$ and 4$\times$ case length. Positive values indicate that the judgment was drawn toward the added side.}
    \label{fig:dose}
\end{figure}

For the addition of a relevant factor, attraction was positive across all models and reasoning levels, indicating a tendency to select the case with added relevant information as determinable even with reasoning enabled. For GPT-5.5 with reasoning, the 95\% CI included zero, but the point estimate remained positive ($0.033$). In contrast, adding irrelevant text produced negative attraction in all five models without reasoning, indicating a tendency to regard the augmented case as undeterminable. Reasoning substantially reduced this effect. Because the length-matched addition averaged only about 8 tokens, we also added irrelevant text to extend cases to approximately twice and four times their original length (Figure~\ref{fig:dose}). Without reasoning, these larger additions amplified the shift toward undeterminable. Reasoning substantially reduced this effect, although statistically significant negative attraction remained in some settings, including Opus 4.5 at both lengths.

LLM judgments shifted in response to input changes despite an unchanged correct answer, indicating a lack of robustness. The shift toward cases with added relevant information suggests that models use the amount of available relevant information as a cue for determinability. These findings demonstrate factor coverage bias. Its direction and response to reasoning distinguish factor coverage bias from context-length effects.

\section{Discussion}
\label{sec:discussion}

We identify factor coverage bias, the tendency to use the amount of available relevant information as a cue for determinability, as a bias in LLM judgment distinct from context-length effects. By distorting determinability judgments, this bias may undermine LLM trustworthiness and usability through unnecessary clarification or premature conclusions. These risks are particularly relevant to tasks that require robust responses to underspecified inputs, including medicine, coding assistance, legal consultation, and customer support \citep{li_MediQ_2024,fang_ClarifyCodeBench_2026a,vincent_InsufficiencyBench_2026,mass_Conversational_2022}. In medicine, the consequences may be especially serious, ranging from incorrect decisions through premature conclusions \citep{graber_Diagnostic_2005} to patient harm from unnecessary invasive testing \citep{ganguli_Cascades_2019} and burdensome questioning. Trustworthy LLMs therefore require both knowledge and the reasoning abilities that experts use in practice. Judging determinability is one such ability.

Given the substantial improvement in determinability judgments with reasoning, allocating inference resources to these judgments is an important design consideration. However, factor coverage bias persisted with reasoning, motivating further training and evaluation using accurately labeled determinable and undeterminable inputs. Such labels are difficult to obtain because experts do not necessarily document their judgments of information sufficiency during information gathering, even when the final conclusion is recorded. Our work addresses this challenge by synthesizing cases with determinability labels, providing an evaluation framework that could be extended to other domains for developing trustworthy LLMs.

\section{Conclusion}
\label{sec:conclusion}
We defined and demonstrated factor coverage bias, the tendency to use the amount of available decision-relevant information as a cue for determinability. This bias may undermine LLM trustworthiness under underspecification through premature conclusions and excessive information gathering. Its direction and response to reasoning distinguish it from context-length effects, informing future efforts to develop trustworthy LLMs. We release a controlled pairwise benchmark for evaluating this bias, providing an evaluation framework for developing trustworthy LLMs.

\section*{Limitations}
\label{sec:limitation}
This study has six limitations. First, the evaluation is limited to five models and to a single synthetic medical benchmark generated by concatenating text templates from clinical scoring systems with explicit decision criteria; it does not include judgments on which clinicians themselves might disagree. In addition, the text chunks were verified by a single physician. Because this verification consists of checking each chunk against explicit item definitions rather than exercising clinical judgment, we expect little inter-rater variability, but we did not measure it.

Second, our evaluation covers only relatively short contexts. Although reasoning substantially reduced negative attraction from irrelevant text, we did not test whether this reduction holds at substantially greater context lengths. The irrelevant-text conditions were designed to distinguish factor coverage bias from context-length effects within our benchmark, rather than to characterize long-context behavior comprehensively.

Third, the pairwise design was chosen to observe the bias while suppressing the influence of each model's abstention tendency; whether the same bias appears in pointwise judgments of a single case has not been tested. Nor have we tested settings closer to actual use, in which information accumulates sequentially through history taking or retrieval-augmented generation; whether the bias manifests there as premature conclusion of information gathering is a question that directly corresponds to the motivation of this study but remains open.

Fourth, this study targets whether models can make the judgment autonomously, using a single zero-shot template that does not supply the reasoning procedure an expert would follow. Prior work has shown that providing the judgment procedure through few-shot prompting improves the accuracy of determinability judgment \citep{watanabe_ClinDetBench_2026}, and the present bias would presumably be mitigated as well; however, because our question is whether LLMs apply such an expert procedure on their own, mitigation by externally supplying the procedure is outside the scope of this study.

Fifth, the amount of relevant information could not be varied: adding further relevant factors to the undeterminable case eventually makes both cases of a pair determinable, so the relevant condition was restricted to a single factor. Even a single factor, however, produced a shift comparable to that of irrelevant text (Table~\ref{tab:relevant_vs_irrelevant}).

Finally, for GPT-5.5 with reasoning enabled, the error rates were near the floor (0.017--0.050); because attraction is defined as a difference in error rates, the range of values it can take is compressed in this setting. The point estimate was positive, as for the other models, but the confidence interval included 0, and we cannot distinguish whether this reflects the absence of the bias or low statistical power due to the small number of errors.

\section*{Ethical Considerations}
Our benchmark consists exclusively of synthetic cases constructed from publicly available scoring criteria and does not include any patient information or private clinical data.
This benchmark is intended solely for research and evaluation purposes and is not designed for direct use in clinical practice or patient-facing applications. The proposed framework does not replace clinical judgment or human supervision.
To ensure transparency and reproducibility, the dataset is publicly released under the MIT License.


\bibliography{references}

\appendix
% Appendix (2026-09-12 v9). The text of the four sections comes first; all
% tables and figures are full width and are flushed as float pages, in number
% order, by the \clearpage after the tables and the one at the end.
% Needs booktabs, graphicx, tcolorbox, array (via makecell); ulem is optional.

\setlength{\tabcolsep}{4pt}
\makeatletter% float pages: floats at the top, not vertically centred
\setlength{\@dblfptop}{0pt}
\setlength{\@dblfpsep}{8pt plus 2pt}
\setlength{\@dblfpbot}{0pt plus 1fil}
\makeatother

\tcbset{promptbox/.style={
  colback=gray!5, colframe=gray!60, boxrule=0.5pt, arc=2pt,
  colbacktitle=gray!90, coltitle=white, fonttitle=\bfseries,
  left=4pt, right=4pt, top=1pt, bottom=1pt, boxsep=1pt,
  before skip=0pt, after skip=4pt}}
\newcommand{\promptinfo}[1]{\hfill{\mdseries\small #1}}
% \uline: from ulem if \usepackage[normalem]{ulem} is in the preamble, else this fallback.
\makeatletter
\@ifpackageloaded{ulem}{}{%
  \newdimen\pxul@depth \pxul@depth=2.6pt
  \def\pxul@word#1{\setbox\z@\hbox{#1}\hbox{\copy\z@\hskip-\wd\z@
    \vrule height-\pxul@depth depth\dimexpr\pxul@depth+0.4pt\relax width\wd\z@}}
  \def\pxul@space{\leaders\hrule height-\pxul@depth depth\dimexpr\pxul@depth+0.4pt\relax
    \hskip\fontdimen2\font plus\fontdimen3\font minus\fontdimen4\font\relax}
  \def\pxul@scan#1 #2\@nil{\pxul@word{#1}\ifx\relax#2\relax\else\pxul@space\pxul@scan#2\@nil\fi}
  \newcommand{\uline}[1]{\pxul@scan#1 \@nil}}
\makeatother

\clearpage
\setcounter{table}{0}
\renewcommand{\thetable}{A.\arabic{table}}
\setcounter{figure}{0}
\renewcommand{\thefigure}{A.\arabic{figure}}

\section{Additional Results}
\label{sec:app_results}

\begin{table*}[t]
  \centering
  \footnotesize
  \caption{Error rate for every model, reasoning level, and condition. Det/Und denote the determinable/undeterminable case of each pair to which the text is added. Matched, 2$\times$, and 4$\times$ are the amounts of irrelevant text (Table~\ref{tab:a7_condition_stats}). The rows reported as reasoning \texttt{on} in the main text are \texttt{max} for GPT-5.5, \texttt{high} for Opus 4.5, and \texttt{on} for Qwen; \texttt{none} denotes no reasoning (thinking off for Qwen).}
  \label{tab:a9_error_rate_all}
  \begin{tabular}{llccccccccc}
    \toprule
    & & & \multicolumn{2}{c}{Add relevant factor} & \multicolumn{6}{c}{Add irrelevant text} \\
    \cmidrule(lr){4-5} \cmidrule(lr){6-11}
    & & & & & \multicolumn{2}{c}{matched} & \multicolumn{2}{c}{2$\times$} & \multicolumn{2}{c}{4$\times$} \\
    \cmidrule(lr){6-7} \cmidrule(lr){8-9} \cmidrule(lr){10-11}
    Model & Reasoning & Baseline & Det & Und & Det & Und & Det & Und & Det & Und \\
    \midrule
    GPT-5.5         & none   & 0.342 & 0.233 & 0.525 & 0.475 & 0.267 & 0.542 & 0.258 & 0.500 & 0.267 \\
                    & low    & 0.067 & 0.017 & 0.100 & 0.067 & 0.042 & 0.058 & 0.058 & 0.058 & 0.083 \\
                    & high   & 0.033 & 0.033 & 0.058 & 0.033 & 0.025 & 0.025 & 0.050 & 0.033 & 0.050 \\
                    & max    & 0.017 & 0.017 & 0.050 & 0.025 & 0.025 & 0.025 & 0.025 & 0.025 & 0.025 \\
    \midrule
    Opus 4.5        & none   & 0.425 & 0.358 & 0.525 & 0.558 & 0.392 & 0.575 & 0.383 & 0.658 & 0.317 \\
                    & low    & 0.225 & 0.083 & 0.417 & 0.283 & 0.217 & 0.325 & 0.217 & 0.358 & 0.183 \\
                    & medium & 0.108 & 0.092 & 0.267 & 0.200 & 0.133 & 0.200 & 0.100 & 0.208 & 0.108 \\
                    & high   & 0.125 & 0.042 & 0.233 & 0.142 & 0.133 & 0.167 & 0.100 & 0.175 & 0.092 \\
    \midrule
    Qwen3.5-27B     & none   & 0.342 & 0.225 & 0.492 & 0.542 & 0.183 & 0.750 & 0.133 & 0.783 & 0.100 \\
                    & on     & 0.100 & 0.050 & 0.242 & 0.208 & 0.108 & 0.217 & 0.133 & 0.208 & 0.200 \\
    \midrule
    Qwen3.5-9B      & none   & 0.375 & 0.350 & 0.433 & 0.617 & 0.275 & 0.733 & 0.183 & 0.800 & 0.133 \\
                    & on     & 0.167 & 0.100 & 0.233 & 0.192 & 0.217 & 0.242 & 0.217 & 0.233 & 0.233 \\
    \midrule
    Qwen3.6-35B-A3B & none   & 0.375 & 0.367 & 0.508 & 0.500 & 0.308 & 0.558 & 0.225 & 0.767 & 0.200 \\
                    & on     & 0.200 & 0.108 & 0.292 & 0.233 & 0.192 & 0.300 & 0.217 & 0.292 & 0.217 \\
    \bottomrule
  \end{tabular}
\end{table*}

\begin{figure*}[t]
    \centering
    \includegraphics[width=\textwidth]{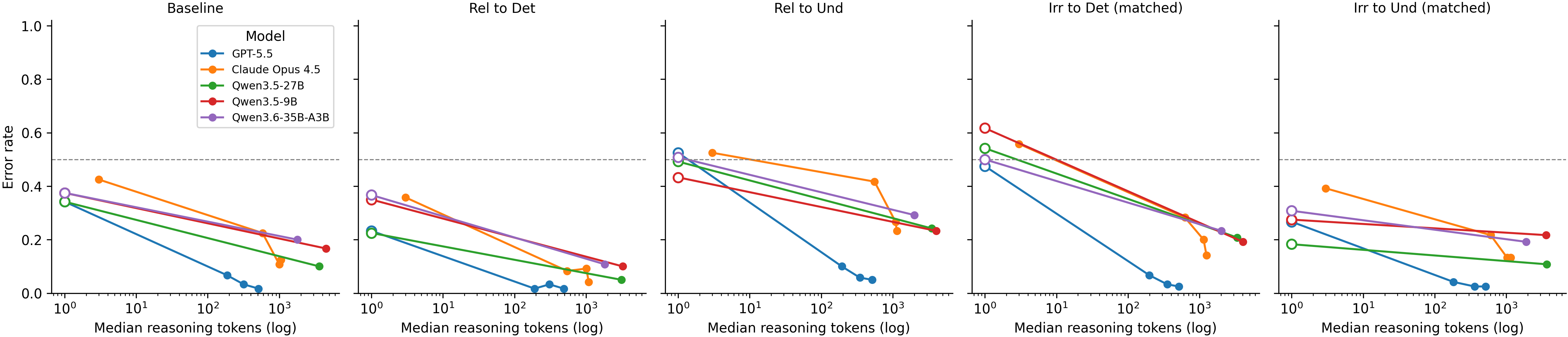}
    \caption{Error rate versus the number of reasoning tokens per run, by model and condition. GPT-5.5 and Opus 4.5 have four reasoning levels and the Qwen models two (thinking off/on). For Opus 4.5, reasoning tokens are estimated as output tokens minus the visible answer because the API does not report them separately, so its \texttt{none} level is not at zero.}
    \label{fig:scaling_reasoning_tokens}
\end{figure*}

Table~\ref{tab:a9_error_rate_all} reports the error rate for every model and reasoning level under all nine conditions, including the intermediate levels of GPT-5.5 and Opus 4.5 and the 2$\times$ and 4$\times$ conditions, and Figure~\ref{fig:scaling_reasoning_tokens} plots each run against its median number of reasoning tokens. Within each model the error rate fell roughly monotonically as the reasoning level, and with it the number of reasoning tokens, increased. The only cells in which the highest level did not lower the error rate are the conditions in which 2$\times$ or 4$\times$ irrelevant text is added to the undeterminable case, for the Qwen models, where the avoidance of the side carrying irrelevant text happens to favor the correct answer without reasoning and reasoning removes it. Across models, however, more reasoning tokens did not mean fewer errors: at their highest levels the three Qwen models consumed more reasoning tokens than GPT-5.5 and Opus 4.5, yet reduced their baseline error rate less (from 0.342--0.375 to 0.100--0.200, against 0.342 to 0.017 for GPT-5.5 and 0.425 to 0.125 for Opus 4.5). The gain from reasoning is therefore bounded by the model rather than by the amount of reasoning tokens.
% Exceptions to "roughly monotonically" (all within 0.02): Opus 4.5 baseline
% medium 0.108 -> high 0.125, Rel to Det low 0.083 -> medium 0.092; GPT-5.5
% Rel to Det low 0.017 -> high 0.033.

\section{Clinical Scoring Systems}
\label{sec:app_scoring_systems}
\begin{table*}[t]
  \centering
  \footnotesize
  \caption{The 15 clinical scoring systems used. Items is the number of scored items and Range the range of possible totals; Rule is the threshold and direction used to generate the ground-truth labels; the decision target is the wording embedded in the prompt. Ref.\ gives the original publication of each system together with a current clinical guideline in which it is used.}
  \label{tab:a2_scoring_systems}
  \begin{tabular}{lcccp{4.0cm}p{3.4cm}}
    \toprule
    Scoring system & Items & Range & Rule & Decision target (as used in the prompt) & Ref. \\
    \midrule
    Alvarado Score            & 8 & 0--10 & $\ge 7$  & acute appendicitis is suggested & \citealp{alvarado_Practical_1986,disaverio_Diagnosis_2020} \\
    Apgar Score               & 5 & 0--10 & $\ge 7$  & the newborn's condition is normal & \citealp{apgar_Proposal_1953,committee_Apgar_2015} \\
    BISAP Score               & 5 & 0--5  & $\ge 3$  & the patient is at higher risk of mortality & \citealp{wu_Early_2008,tenner_American_2013} \\
    Bishop Score              & 5 & 0--13 & $\le 7$  & the cervix is unfavorable for induction & \citealp{bishop_PELVIC_1964,acog_Induction_2009} \\
    Blatchford Score          & 8 & 0--23 & $\ge 1$  & the patient is at high risk and needs intervention & \citealp{blatchford_Risk_2000,_Overview_2012} \\
    CHA$_2$DS$_2$-VASc Score  & 7 & 0--9  & $\ge 2$  & the patient is at high stroke risk and needs anticoagulation & \citealp{lip_Refining_2010,hindricks_2020_2021} \\
    CHADS$_2$ Score           & 5 & 0--6  & $\ge 2$  & the patient is at high stroke risk and needs anticoagulation & \citealp{gage_Validation_2001,takase_JCS_2024} \\
    CURB-65 Score             & 5 & 0--5  & $\ge 3$  & the pneumonia is severe and warrants hospitalization & \citealp{lim_Defining_2003,metlay_Diagnosis_2019} \\
    Centor Score              & 5 & 0--5  & $\le 2$  & no further testing or antibiotics are needed & \citealp{centor_Diagnosis_1981,shulman_Clinical_2012} \\
    Child-Pugh Classification & 5 & 0--15 & $\ge 10$ & the cirrhosis is class C & \citealp{pugh_Transection_1973,europeanassociationforthestudyoftheliver_EASL_2018} \\
    Glasgow-Imrie Score       & 8 & 0--8  & $\ge 3$  & this is severe acute pancreatitis & \citealp{imrie_Singlecentre_1978,blamey_Prognostic_1984,tenner_American_2013} \\
    HAS-BLED Score            & 9 & 0--9  & $\ge 3$  & the patient is at high bleeding risk & \citealp{pisters_Novel_2010,hindricks_2020_2021} \\
    Khorana Score             & 5 & 0--6  & $\ge 3$  & the patient is at high risk of venous thromboembolism & \citealp{khorana_Development_2008,key_Venous_2023} \\
    ORBIT Bleeding Score      & 5 & 0--7  & $\ge 4$  & the patient is at high risk of major bleeding & \citealp{obrien_ORBIT_2015,hindricks_2020_2021} \\
    SIRS Criteria             & 4 & 0--4  & $\ge 2$  & the patient meets the criteria for SIRS & \citealp{bone_Definitions_1992,tenner_American_2013} \\
    \bottomrule
  \end{tabular}
\end{table*}

Table~\ref{tab:a2_scoring_systems} lists the 15 scoring systems with the rule used to generate the ground-truth labels and the decision target embedded in the prompt. For the two systems whose rule is an upper bound (Bishop and Centor), the definition of Section~\ref{sec:method} applies with the inequalities reversed.

\section{Dataset Details}
\label{sec:app_dataset}
\label{sec:app_case_construction}\label{sec:app_baseline}\label{sec:app_irr}% former subsections

Table~\ref{tab:a4_scored_item_chunks} lists the scored-item chunks for the CHADS$_2$ Score, from which the case presentations are built (Section~\ref{sec:method}), and Table~\ref{tab:a8_irrelevant_chunks} the irrelevant text for the same system; one fixed irrelevant text was written per scoring system and amount ($15 \times 3 = 45$ texts), the three amounts are nested, and the same text is added whichever side receives it. Table~\ref{tab:a5_baseline_by_score} gives the composition of the 120 baseline pairs by scoring system, and Table~\ref{tab:a7_condition_stats} reports the measured amount of text added in each condition and the resulting prompt length. Figures~\ref{fig:a3_prompt_examples_1} and~\ref{fig:a4_prompt_examples_2} (Appendix~\ref{sec:app_prompt}) show one pair built from these chunks under the baseline condition and the four conditions in which text is added to the undeterminable side.

\begin{table*}[t]
  \footnotesize
  \begin{minipage}[t]{0.47\textwidth}
  \centering
    \caption{Examples of pre-written scored-item chunks for CHADS$_2$. Positive chunks indicate that the scoring criterion is met, while negative chunks indicate that it is not. Missing item values are represented by omitting the corresponding chunks.}
    \label{tab:a4_scored_item_chunks}
    \begin{tabular}{@{}>{\raggedright\arraybackslash}p{0.24\linewidth}>{\raggedright\arraybackslash}p{\dimexpr0.38\linewidth-8pt}>{\raggedright\arraybackslash}p{\dimexpr0.38\linewidth-8pt}@{}}
      \toprule
      Item & Positive chunk & Negative chunk \\
      \midrule
      Congestive heart failure
        & a history of congestive heart failure
        & no history of congestive heart failure \\
      Hypertension
        & a history of hypertension
        & no history of hypertension \\
      Age $\ge 75$
        & An 80-year-old
        & A 68-year-old \\
      Diabetes mellitus
        & a history of diabetes mellitus
        & no history of diabetes mellitus \\
      Prior stroke or TIA
        & a prior stroke
        & no prior stroke or TIA \\
      \bottomrule
    \end{tabular}
  \end{minipage}\hfill
  \begin{minipage}[t]{0.50\textwidth}
  \centering
    \caption{Examples of pre-written irrelevant text for CHADS$_2$. These sentences do not correspond to any scoring item. Each condition includes the text from the preceding condition plus the sentences shown in its row. Sentence and token counts are cumulative.}
    \label{tab:a8_irrelevant_chunks}
    \begin{tabular}{@{}>{\raggedright\arraybackslash}p{1.7cm}>{\raggedright\arraybackslash}p{\dimexpr\linewidth-1.7cm-8pt\relax}@{}}
      \toprule
      Amount & Sentences added \\
      \midrule
      matched \newline 1 sentence \newline 7 tokens & Family history included maternal osteoarthritis. \\
      \midrule
      2$\times$ \newline 2 sentences \newline 33 tokens & $+$ The patient reported a severe allergy to macrolide antibiotics, which historically caused significant gastrointestinal upset and a diffuse skin rash during childhood. \\
      \midrule
      4$\times$ \newline 5 sentences \newline 98 tokens & $+$ According to the medical record, the patient took oral levothyroxine each morning to manage a long-standing history of uncomplicated primary hypothyroidism. Current regular medications included a daily oral zinc tablet and a nonprescription antihistamine taken during the spring months for seasonal allergic rhinitis. The patient had a penicillin allergy. \\
      \bottomrule
    \end{tabular}
  \end{minipage}
\end{table*}

\begin{table*}[t]
  \centering
  \footnotesize
  \caption{Composition of the 120 baseline pairs by scoring system: eight pairs per system, four with the correct answer A and four with B. Missing = number of missing scored items per case, median [range]. Det margin = distance of the determinable side from the threshold (0 = exactly at it); Und margin = points the undeterminable side would need to move to become determinable. Tokens are mean (SD) per case.}
  \label{tab:a5_baseline_by_score}
  \begin{tabular}{lcccrr}
    \toprule
    Scoring system & Missing & Det margin & Und margin & Det tokens & Und tokens \\
    \midrule
    Alvarado Score            & 4 [2--5] & 0 [0--1] & 1 [1--3] & 48.2 (8.6)  & 51.6 (10.4) \\
    Apgar Score               & 3 [2--3] & 0 [0--0] & 2 [1--3] & 23.5 (5.0)  & 27.8 (5.3) \\
    BISAP Score               & 2 [2--2] & 0 [0--0] & 1 [1--1] & 32.8 (3.4)  & 32.9 (2.4) \\
    Bishop Score              & 2 [2--2] & 0 [0--1] & 1 [1--3] & 40.2 (2.9)  & 39.9 (2.3) \\
    Blatchford Score          & 4 [2--5] & 3 [0--9] & 1 [1--1] & 44.6 (8.0)  & 40.6 (8.3) \\
    CHA$_2$DS$_2$-VASc Score  & 2 [2--5] & 1 [0--4] & 1 [1--2] & 41.6 (8.2)  & 45.4 (9.9) \\
    CHADS$_2$ Score           & 3 [2--4] & 0 [0--1] & 1 [1--2] & 32.2 (4.2)  & 32.9 (3.8) \\
    CURB-65 Score             & 2 [2--2] & 0 [0--0] & 1 [1--1] & 52.5 (3.4)  & 50.1 (1.9) \\
    Centor Score              & 2 [2--3] & 0 [0--1] & 1 [1--2] & 26.9 (3.8)  & 26.8 (5.1) \\
    Child-Pugh Classification & 2 [2--2] & 0 [0--1] & 2 [1--2] & 44.2 (6.4)  & 42.5 (5.3) \\
    Glasgow-Imrie Score       & 3 [2--4] & 0 [0--2] & 1 [1--2] & 66.9 (9.7)  & 64.6 (8.0) \\
    HAS-BLED Score            & 4 [3--6] & 0 [0--1] & 1 [1--2] & 40.6 (8.5)  & 40.8 (7.6) \\
    Khorana Score             & 2 [2--3] & 0 [0--1] & 1 [1--2] & 44.8 (5.4)  & 47.2 (6.8) \\
    ORBIT Bleeding Score      & 2 [2--3] & 0 [0--1] & 2 [1--3] & 31.9 (4.8)  & 30.4 (3.7) \\
    SIRS Criteria             & 2 [2--2] & 0 [0--0] & 1 [1--1] & 39.4 (3.1)  & 37.9 (1.6) \\
    \midrule
    All (120 pairs)           & 2 [2--6] & 0 [0--9] & 1 [1--3] & 40.7 (11.9) & 40.8 (11.4) \\
    \bottomrule
  \end{tabular}
\end{table*}

\begin{table*}[t]
  \centering
  \footnotesize
  \caption{Amount of text added per case in each condition. Rel/Irr: addition of a relevant factor/irrelevant text; Det/Und: the determinable/undeterminable case receiving it (Irr rows apply to either side). Added tokens count only the inserted text; prompt tokens are the mean length of the complete prompt actually sent to the model, i.e., the instruction template plus the case description and any added text. Both columns report means. Token counts use tiktoken (\texttt{o200k\_base}); baseline case descriptions average 41 tokens.}
  \label{tab:a7_condition_stats}
  \begin{tabular}{lrr}
    \toprule
    Condition & Added tokens & Prompt tokens \\
    \midrule
    Baseline          & ---  & 140 \\
    Rel to Det        & 8    & 147 \\
    Rel to Und        & 8    & 148 \\
    Irr (matched)     & 8    & 148 \\
    Irr (2$\times$)   & 43   & 182 \\
    Irr (4$\times$)   & 127  & 267 \\
    \bottomrule
  \end{tabular}
\end{table*}

\section{Prompt Template and Examples}
\label{sec:app_prompt}

Figure~\ref{fig:a1_prompt_template} shows the template, identical for every instance and every model; it names the scoring system but does not list its items or threshold, and no judgment procedure is supplied (Section~\ref{sec:experiments}). Figures~\ref{fig:a3_prompt_examples_1} and~\ref{fig:a4_prompt_examples_2} show the prompts actually sent for one pair of the CHADS$_2$ Score under the baseline condition and the four conditions in which text is added to the undeterminable side; the conditions that add text to the determinable case add the same text to Case A instead, and text added relative to the baseline is underlined.

\clearpage% flush the tables first, so that Figure A.2 stays behind them

\begin{figure*}[t]
\begin{tcolorbox}[promptbox, title={Prompt template}]
\small\ttfamily\raggedright
Two cases are presented below.\\[0.3em]
Case A:\\
\{case\_A\}\\[0.3em]
Case B:\\
\{case\_B\}\\[0.3em]
Based on the \{score\_name\}, which case has sufficient information to determine whether \{decision\_target\}?\\
Answer with a single letter, "A" or "B". Output only that letter and nothing else.
\end{tcolorbox}
\caption{The prompt template, identical for every instance and every model, sent as a single user message with no system prompt.}
\label{fig:a1_prompt_template}
\end{figure*}

\begin{figure*}[!t]

\begin{tcolorbox}[promptbox, title={Baseline\promptinfo{tokens: Case A 39, Case B 41, prompt 142}}]
\small\ttfamily\raggedright
Two cases are presented below.\\[0.3em]
Case A:\\
An 80-year-old patient with atrial fibrillation was evaluated for stroke risk to guide anticoagulation. The patient had a history of congestive heart failure and no history of diabetes mellitus.\\[0.3em]
Case B:\\
A patient with atrial fibrillation was evaluated for stroke risk to guide anticoagulation. The patient had no history of congestive heart failure, a history of hypertension, and no history of diabetes mellitus.\\[0.3em]
Based on the CHADS2 Score, which case has sufficient information to determine whether the patient is at high stroke risk and needs anticoagulation?\\
Answer with a single letter, "A" or "B". Output only that letter and nothing else.
\end{tcolorbox}

\begin{tcolorbox}[promptbox, title={Rel to Und\promptinfo{tokens: Case A 39, Case B 45, prompt 146}}]
\small\ttfamily\raggedright
Two cases are presented below.\\[0.3em]
Case A:\\
An 80-year-old patient with atrial fibrillation was evaluated for stroke risk to guide anticoagulation. The patient had a history of congestive heart failure and no history of diabetes mellitus.\\[0.3em]
Case B:\\
A \uline{68-year-old} patient with atrial fibrillation was evaluated for stroke risk to guide anticoagulation. The patient had no history of congestive heart failure, a history of hypertension, and no history of diabetes mellitus.\\[0.3em]
Based on the CHADS2 Score, which case has sufficient information to determine whether the patient is at high stroke risk and needs anticoagulation?\\
Answer with a single letter, "A" or "B". Output only that letter and nothing else.
\end{tcolorbox}

\begin{tcolorbox}[promptbox, title={Irr to Und (matched)\promptinfo{tokens: Case A 39, Case B 48, prompt 149}}]
\small\ttfamily\raggedright
Two cases are presented below.\\[0.3em]
Case A:\\
An 80-year-old patient with atrial fibrillation was evaluated for stroke risk to guide anticoagulation. The patient had a history of congestive heart failure and no history of diabetes mellitus.\\[0.3em]
Case B:\\
A patient with atrial fibrillation was evaluated for stroke risk to guide anticoagulation. The patient had no history of congestive heart failure, a history of hypertension, and no history of diabetes mellitus. \uline{Family history included maternal osteoarthritis.}\\[0.3em]
Based on the CHADS2 Score, which case has sufficient information to determine whether the patient is at high stroke risk and needs anticoagulation?\\
Answer with a single letter, "A" or "B". Output only that letter and nothing else.
\end{tcolorbox}

\caption{The prompts actually sent for one pair (CHADS$_2$ Score) under Baseline, Rel to Und, and Irr to Und (matched); Figure~\ref{fig:a4_prompt_examples_2} continues with 2$\times$ and 4$\times$. Case A, the determinable side and correct answer, is identical throughout; the underlined text is added to Case B: a missing scored item (age) in Rel to Und, and one, two, or five irrelevant sentences (Table~\ref{tab:a8_irrelevant_chunks}) in Irr to Und. Rel to Det and Irr to Det add the text to Case A instead. Title bars give the token counts (tiktoken \texttt{o200k\_base}).}
\label{fig:a3_prompt_examples_1}
\end{figure*}

\begin{figure*}[!t]

\begin{tcolorbox}[promptbox, title={Irr to Und (2$\times$)\promptinfo{tokens: Case A 39, Case B 74, prompt 175}}]
\small\ttfamily\raggedright
Two cases are presented below.\\[0.3em]
Case A:\\
An 80-year-old patient with atrial fibrillation was evaluated for stroke risk to guide anticoagulation. The patient had a history of congestive heart failure and no history of diabetes mellitus.\\[0.3em]
Case B:\\
A patient with atrial fibrillation was evaluated for stroke risk to guide anticoagulation. The patient had no history of congestive heart failure, a history of hypertension, and no history of diabetes mellitus. \uline{Family history included maternal osteoarthritis. The patient reported a severe allergy to macrolide antibiotics, which historically caused significant gastrointestinal upset and a diffuse skin rash during childhood.}\\[0.3em]
Based on the CHADS2 Score, which case has sufficient information to determine whether the patient is at high stroke risk and needs anticoagulation?\\
Answer with a single letter, "A" or "B". Output only that letter and nothing else.
\end{tcolorbox}

\begin{tcolorbox}[promptbox, title={Irr to Und (4$\times$)\promptinfo{tokens: Case A 39, Case B 139, prompt 240}}]
\small\ttfamily\raggedright
Two cases are presented below.\\[0.3em]
Case A:\\
An 80-year-old patient with atrial fibrillation was evaluated for stroke risk to guide anticoagulation. The patient had a history of congestive heart failure and no history of diabetes mellitus.\\[0.3em]
Case B:\\
A patient with atrial fibrillation was evaluated for stroke risk to guide anticoagulation. The patient had no history of congestive heart failure, a history of hypertension, and no history of diabetes mellitus. \uline{Family history included maternal osteoarthritis. The patient reported a severe allergy to macrolide antibiotics, which historically caused significant gastrointestinal upset and a diffuse skin rash during childhood. According to the medical record, the patient took oral levothyroxine each morning to manage a long-standing history of uncomplicated primary hypothyroidism. Current regular medications included a daily oral zinc tablet and a nonprescription antihistamine taken during the spring months for seasonal allergic rhinitis. The patient had a penicillin allergy.}\\[0.3em]
Based on the CHADS2 Score, which case has sufficient information to determine whether the patient is at high stroke risk and needs anticoagulation?\\
Answer with a single letter, "A" or "B". Output only that letter and nothing else.
\end{tcolorbox}

\caption{The prompts for the same pair under Irr to Und with the 2$\times$ and 4$\times$ amounts of irrelevant text (continued from Figure~\ref{fig:a3_prompt_examples_1}). The irrelevant text is nested: each amount extends the preceding one (Table~\ref{tab:a8_irrelevant_chunks}). Text added relative to the baseline is underlined.}
\label{fig:a4_prompt_examples_2}
\end{figure*}

\clearpage

\end{document}